# How We Define Harm Impacts Data Annotations: Explaining How Annotators Distinguish Hateful, Offensive, and Toxic Comments

Angela Schöpke-Gonzalez*, Siqi Wu*, Sagar Kumar**, Paul J. Resnick*, Libby Hemphill*

*University of Michigan School of Information, **Northeastern University*

## Abstract

Computational social science research has made advances in machine learning and natural language processing that support content moderators in detecting harmful content. These advances often rely on training datasets annotated by crowdworkers for harmful content. In designing instructions for annotation tasks to generate training data for these algorithms, researchers often treat the harm concepts that we train algorithms to detect – "hateful", "offensive", "toxic", "racist", "sexist", etc. – as interchangeable. In this work, we studied whether the way that researchers define "harm" affects annotation outcomes. Using Venn diagrams, information gain comparisons, and content analyses, we reveal that annotators do not use the concepts "hateful", "offensive", and "toxic" interchangeably. We identify that features of harm definitions and annotators' individual characteristics explain much of how annotators use these terms differently. Our results offer empirical evidence discouraging the common practice of using harm concepts interchangeably in content moderation research. Instead, researchers should make specific choices about which harm concepts to analyze based on their research goals. Recognizing that researchers are often resource constrained, we also encourage researchers to provide information to bound their findings when their concepts of interest differ from concepts that off-the-shelf harmful content detection algorithms identify. Finally, we encourage algorithm providers to ensure their instruments can adapt to contextually-specific content detection goals (e.g., soliciting instrument users' feedback).

# 1 Introduction

The volume and velocity of content generated by social media outpaces human moderators' abilities to keep up. Human moderators alone cannot review all content, and some extent of automated or computational assistance can support content moderation. However, recent research identifies a number of challenges for automated content moderation in detecting harmful content, such as obscuring the politics of moderation decision making (Gorwa et al., 2020), reliability of automated moderation (Ganesh & Bright, 2020), and tensions between encryption and moderation (Gillespie et al., 2023). Supervised machine learning is the most common approach for automated detection, relying on data annotated by human to train algorithms to recognize classes of harmful content.

In designing instructions for the annotation tasks to generate training data, researchers have often treated the harm concepts we train the algorithms to detect – "hateful", "offensive", "toxic", "racist", "sexist", etc. – as interchangeable (D'Sa et al., 2020). However, research has also pointed out that there may be meaningful distinctions between these concepts (Waseem et al., 2017). These distinctions can impact how the human annotators identify what counts as harmful content, and consequently the data that harmful content detection models are trained on. These models' outputs can then disproportionately harm minoritized populations by motivating increased flagging and removal of these groups' content, for example (Haimson et al., 2021; Sap et al., 2019). In order to address the potential effects of relative inattention to concept definition in annotation tasks, this study investigates:

> RQ 1: Do annotators use the labels *hateful*, *offensive*, and *toxic* interchangeably?
>
> RQ 2: If not, what features of definitions or comments explain the differences between *hateful*, *offensive*, and *toxic* concepts?

Building on prior literature and using a dataset of social media comments from Reddit, Twitter, and YouTube annotated by MTurk workers as hateful, offensive, or toxic (Wu et al., 2023), we developed a codebook that captures the distinguishing and overlapping features of the three concepts. We applied this codebook to a subset of Wu et al.'s dataset using content analysis techniques, and then performed statistical analysis to assess the degree to which these features explain whether a comment was labeled hateful, offensive, or toxic.

We found that annotators interpret hateful, offensive, and toxic as distinct concepts. Some of these distinctions may be explained by concept definitions themselves, and some may be explained by annotators' individual characteristics. While it is perhaps intuitive that these concepts differ, our results characterize the potential effects that an annotation task's definition of harm can have on the annotation outcomes. We discourage the common practice of using these harm terms interchangeably in content moderation research, since each concept provides distinct information about different types of harm. Instead, we encourage researchers to make intentional and specific choices about how to define harm in line with their research goals, and to ensure that their annotation task design provides annotators with the information necessary to complete the task requested. We also encourage researchers to appropriately contextualize their findings when off-the-shelf algorithms' definitions of harm differ from the researchers' conception of harm. Finally, we encourage harmful content detection algorithm providers to consider how to ensure that their instruments can adapt to contextually-specific harmful content detection goals (e.g., soliciting instrument users' feedback). In sum, our work shows that how concept definitions in the annotation tasks impacts the way that annotators interpret and label harm, and consequently how the algorithms trained on those labels identify harm.

## 2 Background & Literature Review

In designing algorithms to support content moderation, researchers often discuss the concepts we train these algorithms to detect – "hate", "offensive", "toxic", "terroristic", "obscene", "abusive" and "uncivil" – as interchangeable (D'Sa et al., 2020). We argue that the relative inattention to understanding the nuanced differences in how we define these concepts could impact training data, which in turn contribute to the model outputs that may disproportionately harm historically marginalized populations by overflagging their content or missing content that harms them (Duguay et al., 2020). We review these concepts, how prior work has defined them, and notable features that characterize these definitions. We summarize this review in Table 1. Throughout this section, when we cite a particular definition, the full-text of this definition can be found in Table 1.

### *2.1 Hate Speech*

"Hate speech" is a commonly used phrase to describe harmful content online, but finds a fairly wide interpretation across scholarship. For example, Davidson et al.'s (2017, p. 1) definition of hate speech explicitly focuses on expressions *directed at* groups or members of a group and that which the comment author *intended* to be "derogatory, to humiliate, or to insult". Nockleby by way of Schmidt and Weigand (2017, p. 1) in turn suggests that hate speech can also encompass expressions *directed at* individual persons or groups, and comments with particular *effects* on their recipients (i.e., "disparages"). ElSherief, Kulkarni, et al. (2018, p. 1) and ElSherief, Nilizadeh, et al. (2018, p. 52) specify that hate speech only encompasses expressions *directed at* an individual person that have particular *effects* on that person (i.e., "denigrates"). Each of these definitions implies *directedness* at an entity, but the entity's nature varies across definitions. One definition refers to author *intent* and two refer to the *effect* that a hateful comment has on its recipient.

Beyond *directedness*, author *intent*, and *effects* on a comment recipient, definitions describe particular *behaviors* that characterize hate speech. Waseem and Hovy's (2016, p. 89) description of *behaviors* expressed in hate speech are perhaps the most extensive and specific of those definitions we examine here. Their definition clearly defines specific *behaviors* – use of a slur, attacking, seeking to silence, criticizing, misrepresenting, etc. – that when used in combination with *directedness* toward an individual or group and, in some cases, with particular author *intents*, manifest hate speech.

## 2.2 Offensive

In efforts to identify hate speech online, Hatebase.org – a collaboration between Dark Data Project (*About*, 2023) and The Sentinel Project (*Home*, 2018) – collected a now popularly used lexicon of *offensive* language that social media platform users felt often characterized hate speech (*Hatebase*, n.d.). However, when Davidson et al. (2017) investigated differences between hate speech and offensive language, they found that only approximately 5% of content containing offensive language was interpreted as hate speech by human coders. Their findings suggest that "offensive" is not a good proxy for "hate", and the terms share less overlap than Hatebase.org indicates. Wiegand et al. (2018, p. 1) combine these two earlier definitions in their own definition of offensiveness and, like some conceptions of hatefulness, Wiegand et al.'s offensiveness is characterized by its *directedness* at and *effect* on an individual. However, like Hatebase.org's offensiveness, Wiegand et al. characterize offensiveness according to the content of phrases that the comment uses (i.e., "obscene").

## 2.3 Toxic

In addition to popular terms hateful and offensive, many studies use the term *toxicity*. One especially popular operationalization of toxicity comes from moderation support algorithm Perspective API, which in turn uses Wulczyn et al.'s (2017) definition of toxicity. In this definition, toxic differs from hate speech and offensive language in that it does not refer to qualities like *directedness*, *intention*, *behaviors*, or *specific language*, but rather only describes a comment's *effect*. In the context of gaming communities, Beres et al. (2021, p. 1) describe toxicity as an umbrella term for various types of *behaviors* like abuse, harassment, or flaming *directed* at other community members, thus resembling hate speech. However, like Perspective API's definition, Beres et al's toxicity is characterized by its *effects* on gameplay (i.e., participation in the community), differentiating it from definitions of hatefulness.

Other work specifically interested in harmful content in multiplayer online games defines toxic somewhat differently: "the use of profane language by one player to insult or humiliate a different player in his own team" (Märtens et al., 2015, p. 3). Related, Gilda et al. (2022, p. 360) talk about toxic as a "type of conversation widely found online which are insulting and violent in nature". Both of these definitions approximate a combination of definitions of offensive language and the *directedness* and *intention* component of hate speech definitions.

Toxicity as a concept varies widely in its definitions across scholarship, sometimes resembling hate speech or offensiveness. In other instances, toxicity describes an umbrella concept for different types of harmful content. In still other instances, toxicity refers to a comment's perceived *effects* on participants in an online community.

## 2.4 Additional Terms

In addition to hate speech, offensive language, and toxic content, expressions to describe harmful user-generated content online include "terroristic", "obscene", "abusive", and "uncivil". Gorwa et al. (2020) describe how members of the Global Internet Forum to Counter Terrorism (GIFCT), including Facebook, Google, Twitter, Microsoft, and at least nine other large Internet companies, each have their own policies defining what constitutes "terrorist content". However, these definitions likely shape each other through participating in the GIFCT's collaborative Shared Industry Hash Database (Gorwa et al., 2020). Beyond private definitions of "terrorist content", additional studies use the phrase "obscene" to characterize harmful user-generated content. For example, Rojas-Galeano's (2017, p. 1) definition of "obscene" resembles how Hatebase interprets offensive language, and uses the word "offensive" to characterize obscenity. Additional work talks about "abusive" speech (Papegnies et al., 2018, 2019), which, like hate speech, is characterized by *directedness*, *intent*, or *behavior*. Finally, some studies use the phrase "uncivil" to describe harmful user-generated content (Coe et al., 2014), strongly emphasizing *directedness*, while also referring to the *tone* or *communication style* used in discussion, suggesting context matters more for "uncivil" than for offensive language, for example.

## *2.5 Summary*

In our review of various terms describing harmful content, we found variability in definitions for the same term and across terms, and a series of features along which these definitions vary: directedness, who or what a comment is directed at, effect of a comment on readers, author's intent, type of behavior exhibited by author, presence of specific terms, and tone of a comment (see Table 1 for description of how these definitional features relate to each term as a particular author defines it). Two of these features – author's intent and the effect of a comment on a reader – can be established only by asking authors and readers about their intentions and perceived effects. In our study, we assess whether features of a comment's content and structure can explain differences in annotated datasets' representations of harm.

**Table 1. Review of Harm Concept Definitions and Features**

| Term | Definition | Definitional Features |
|---|---|---|
| Hate | "...language that is used to express hatred towards a targeted group or is intended to be derogatory, to humiliate, or to insult the members of the group" (Davidson et al., 2017, p. 1) | Directedness, who or what a comment is directed at, author intent |
| Hate | Language that: "...uses a sexist or racial slur. attacks a minority. seeks to silence a minority. criticizes a minority (without a well founded argument). promotes, but does not directly use, hate speech or violent crime. criticizes a minority and uses a straw man argument. blatantly misrepresents truth or seeks to distort views on a minority with unfounded claims. shows support of problematic hashtags. E.g. "#BanIslam", "#whoriental", "#whitegenocide" negatively stereotypes a minority. defends xenophobia or sexism. contains a screen name that is offensive, as per the previous criteria, the tweet is ambiguous (at best), and the tweet is on a topic that satisfies any of the above criteria." (Waseem & Hovy, 2016, p. 89) | Directedness, who or what a comment is directed at, author intent, type of behavior exhibited by author |
| Hate | "...denigrates a person because of their innate and protected characteristics" (ElSherief, Kulkarni, et al., 2018, p. 1; ElSherief, Nilizadeh, et al., 2018, p. 52) | Directedness, who or what a comment is directed at, effect of a comment on readers |
| Hate | "...any communication that disparages a person or a group on the basis of some characteristic such as race, color, ethnicity, gender, sexual orientation, nationality, religion, or other characteristic" Nockleby by way of Schmidt and Weigand (2017, p. 1) | Directedness, who or what a comment is directed at, effect of a comment on readers |
| Offensive | Specific words, phrases, or collections of symbols that can collectively be interpreted as hateful (Hatebase, n.d.) | Presence of specific terms |
| Offensive | "hurtful, derogatory or obscene comments made by one person to another person" (Wiegand et al., 2018, p. 1) | Directedness, who or what a comment is directed at, effect of a comment on readers, presence of specific terms |
| Toxic | "a rude, disrespectful, or unreasonable comment that is likely to make you leave a discussion" (Perspective API via Wulczyn et al. 2017) | Effect of a comment on readers |
| Toxic | "various types of negative behaviors involving abusive communications directed towards other players (i.e., harassment, verbal abuse, and flaming) and disruptive gameplay that violates the rules and social norms of the game (i.e., griefing, spamming, and cheating)" (Beres et al., 2021, p. 1) | Directedness, type of behavior exhibited by author, who or what a comment is directed at, effect of a comment on readers |
| Toxic | "the use of profane language by one player to insult or humiliate a different player in his own team" (Märtens et al., 2015) | Directedness, who or what a comment is directed at, author intent, presence of specific terms |
| Toxic | "type of conversation widely found online which are insulting and violent in nature" (Gilda et al., 2022) | Speaker intention, presence of specific terms |
| Terroristic | GIFCT's Shared Industry Hash Database of terms (Gorwa et al., 2020) | Presence of specific terms |
| Obscene | "...the use of rude words or offensive expressions" (Rojas-Galeano, 2017, p. 12:1) | Presence of specific terms |
| Abusive | Speech that demonstrates disrespect toward another person (Papegnies et al., 2017), or speech designed to "personally attack others or discriminate them based on race, religion, or sexual orientation. It can also include more community-specific aspects..." (Papegnies et al., 2019) | Directedness, who or what a comment is directed at, author intent, type of behavior exhibited by author |
| Uncivil | "...features of discussion that convey an unnecessarily disrespectful tone toward the discussion forum, its participants, or its topics" (Coe et al., 2014) | Directedness, who or what a comment is directed at, tone |

## 3 Methods

To understand the annotation variances caused by different concept definitions, we study three prevalently used conceptions of harmful content – hateful (H), offensive (O), and toxic (T). The dataset we analyzed (Wu et al., 2023) includes annotations by MTurk workers identifying which, if any, of these concepts applies to 3481 social media comments posted in response to political news posts and videos on Reddit, Twitter, and YouTube in August, 2021. Each comment received annotations from five MTurk workers. Each worker assessed whether each of the three concepts applied to the comment. See Table 2 for the specific concept definitions used in Wu et al.'s annotation task that generated the dataset.

In most of our analyses, we aggregate the annotations from the five MTurk workers, making a majority vote label for each of the concepts. If a comment received three or more annotations indicating the presence of a concept, the comment received a *true* value for that concept. If it received two or less annotations indicating the presence of a concept, the comment received a *false* value for that concept.

Members of our team manually reviewed comments from the dataset to develop a codebook of comment features and analyzed whether these features explain if social media comments in the dataset received an H, O, or T label from the majority of MTurk workers. Additional details about the dataset and data collection process are available from the archived dataset record.[1]

**Table 2. Definitions of "hateful", "offensive", and "toxic" that were provided to MTurk workers in the instructions about how to annotate the collection of Reddit, Twitter, and YouTube comments.**

| Concept | Definition |
| --- | --- |
| **Hateful** | "expresses hatred **towards a targeted group** or is *intended to be derogatory, to humiliate, or to insult* the **members of the group**" (Davidson et al. 2017) |
| **Offensive** | "contains *hurtful, derogatory*, or **obscene comments**" (Wiegand et al. 2018) |
| **Toxic** | "a *rude, disrespectful, or unreasonable* comment that is *likely to make readers want to leave a discussion*" (Wulczyn et al., 2017) |

*Table Notes: Bold and italicized font types have been added to these definitions for the purpose of this article. Bold text denotes parts of the term definition that specify comment features like directedness or type of content. Italicized text denotes parts of the term definition that specify either author's intent or the effect of a comment on a reader.*

---

[1] https://doi.org/10.3886/45fc-9c8f

## *3.1 Content Analysis*

We performed content analysis (Krippendorff, 2019, p. 24) to identify comment features in the dataset that may explain whether a comment was majority labeled as H, O, or T. We developed our own codebook using a directed coding approach (Hsieh & Shannon, 2005) on a sample of data (220 comments), and applied this codebook to a larger sample (909 comments including the original 220). We then used generalized linear mixed models to characterize patterns among the codes.

### 3.1.1 Codebook Development

Before beginning codebook development, three authors met to discuss how to navigate reviewing content that may cause emotional distress. Strategies included taking breaks and ensuring that we had timespace after reviewing content to externalize anger or hopelessness. We noted differences in our life experiences that could shape how we interpret content. For example, one researcher's relationship with immigration, Afghanistan, and gender-based violence meant that she was particularly attuned to nuances in texts concerning these topics. The second researcher's experiences studying white supremacy and homophobia made her aware of references to religious, racial, and sexual minorities. The third researcher's previous personal and professional work unpacking ideas of "conspiracy" or "misinformation" attuned him to nuances in texts related to these topics.

To develop our codebook, we randomly selected 120 comments that any annotator had indicated were H, O, and/or T. Two of the researchers independently used directed coding (Hsieh & Shannon, 2005) to code the first 60 of the 120 comments, and then met to discuss emergent patterns they noticed. The two researchers then independently reviewed the remaining 60 comments of the random sample. Once the two authors had coded this 120-comment sample, one of the researchers, together with the third researcher, coded an additional 100 randomly selected comments to assess whether the codes developed by the first two researchers could be transferable to a third researcher. They refined and reapplied codes until they reached consensus.

In total, we identified six textual features (see Table 3 for feature descriptions and examples) that may explain differences across HOT-labeled comments: *directedness, pronoun use, behavior, multiple behaviors, communication style,* and *sarcasm*. Some of these features overlap with definitional features reviewed in section 2, specifically: directedness, who or what a comment is directed at, and type of behavior exhibited by the comment author (Beres et al., 2021; Waseem & Hovy, 2016). *Communication style* and *sarcasm* approximate tone as a feature (Coe et al., 2014), and *pronoun use* and *multiple behaviors* are codes that newly emerged from the data. In addition, some of these features overlap with existing codebooks like Goyal et al.'s (2022) specialized rater pools codebook which includes *insult* and *threat* as parent categories describing behaviors. In research concerning ethnic bias of language models, Jeong et al. (2022) also identified targetedness (what we call *directedness*) of a comment at a group or individual as a feature of what characterizes offensiveness in Korean language posts. *Directedness* (as opposed to "generalizedness") of a comment is also a feature in Waseem et How We Define Harm Impacts Data Annotationsal.'s (2017) proposed typology of harmful content assessment. In addition to these features, our codebook encompasses a wider range of features that may explain differences in annotations across H, O, and T labels.

**Table 3. Codebook including descriptions of explanatory features**

| Feature | Description | Codes | Code Description |
|---|---|---|---|
| Directedness | Whether the text is aimed at a particular entity. | Directed | Comments that are aimed at a particular entity. Can be further classified into sub-categories according to the entity at which the comment is aimed: *groups*, *individuals*, or *behaviors*. |
| | | Undirected | Comments that are not aimed at any particular entity. |
| Pronoun Use | The types of pronouns that text authors use in their comments. | 1st person | Comments that use 1st person pronouns (i.e., I). |
| | | 2nd person | Comments that use 2nd person pronouns (i.e., you). |
| | | 3rd person | Comments that use 3rd person pronouns (i.e., he, him, his, himself, she, her, hers, herself, it, its, itself, they, them, their, theirs, and themselves). |
| Behavior | A characterization of the actions that the text authors express via their comments. | Name-calling | Using a specific term to deride a particular entity. |
| | | Insulting | Implicitly deriding a particular entity without using a specific term. |
| | | Accusation | Insinuating that an entity acted wrongfully. |
| | | Threat | Implicitly or explicitly suggesting that physical harm toward a particular entity would be desirable. |
| | | Misinformation | Suggesting that commonly held beliefs are incorrect, when that suggestion could potentially lead to physically harming people. |
| | | Benign | Behaviors that do not appear to have discernible harmful characteristics. |
| Multiple Behaviors | Whether the comment features more than one behavior as per the "Behavior" item. | Yes | Multiple behaviors appear in the comment. |
| | | No | Multiple behaviors do not appear in the comment (i.e., no behaviors or only one behavior appear). |
| Communication Style | Different tones with which a commenter might express themselves. Our codes and code definitions are derived from (*The Four Basic Styles of Communication*, 2014) | Passive | When a commenter avoids expressing their opinions or feelings, protecting their rights, and identifying and meeting their needs. |
| | | Aggressive | When a commenter expresses their feelings and opinions and advocates for their needs in a way that violates the rights of others. |
| | | Passive-Aggressive | When a commenter appears passive on the surface but is really acting out anger in a subtle, indirect, or behind-the-scenes way. |
| | | Assertive | When a commenter clearly states their opinions and feelings, and firmly advocates for their rights and needs without violating the rights of others. |
| Sarcasm | The presence of "the use of remarks that clearly mean the opposite of what they say, made in order to hurt someone's feelings or to criticize something in a humorous way" ("Sarcasm," 2022) | Yes | Sarcasm appears in the comment. |
| | | No | Sarcasm does not appear in the comment. |

## 3.1.2 Statistical Analyses

In order to identify patterns across the six coded features whether these patterns explain differentiation among majority vote H, O, and T annotations, we used generalized linear regression models through R's stats package (R Core Team, 2022). These models can be described by the general equation:

$$\Pr Pr(y = 1) = \theta = logit^{-1}(\beta_0 + \beta_1 x_1 + \beta_2 x_2 + \ldots + \beta_n x_n + \alpha_i)$$

where *y* denotes the binary outcome variable for each concept, *x* denotes each independent variable, and $\alpha_i$ denotes a fixed effect. Since Wu et al. (2023) randomly assigned a group of five annotators from their annotator pool to each social media comment, we needed to account for the effect on labeling of both the individual annotator and the group of five annotators that labeled each comment. To account for the effect of each individual annotator on a group's likelihood to label a comment with each concept, we calculated a fixed effect $\alpha_i$ for each comment *i* by concept; *j* denotes an annotator; *x* denotes a label (0 or 1); *k* denotes a single comment that an annotator has labeled; and *n* denotes the total number of comments an annotator has labeled:

$$\alpha_i = \frac{1}{5}\sum_{j=1}^{5}\left(\frac{1}{n}\sum_{k=1}^{n} x^i_{k,j}\right)$$

Since the features *directed at* and *pronoun use* apply only to directed comments, we separated the dataset into subsamples of directed comments (823 comments total) and undirected comments (87 comments total). Given undirected comments' small sample size, we statistically analyzed only directed comments. Models with the directed comment subset included predictors *directed at, pronoun use, behavior, communication style, sarcasm,* and *rater group effect*.

For regression analyses, we used 11 binary outcome variables (see Table 4 for overview of all outcome variables). Analyses of H (all), O (all), and T (all) outcome variables helped us understand features of an *entire* concept. Analyses of H (only), O (only), and T (only) outcome variables helped us understand features unique to comments labeled as *only one* concept. Subsets featuring overlapping concepts as outcome variables – HO, HT, OT, HOT, or any overlap – helped us identify how comment features intersect between concepts. When including all variables as predictors, most reported models (see Results) offered a statistically significant better fit than null models at a .05 level as per likelihood ratio tests (see Appendix for likelihood ratio test results) except for H (Only), O (Only), HO, HT, which can be expected given the small number of comments with a value of TRUE for each of these models (see Table 4 for count of TRUE values for each outcome variable). In order to understand meaningful differences between H (Only), O (Only), HO, HT, and other concepts through statistically significant models, we performed variable selection based on theory and observed variable significance for these outcome variables. The resulting models for binary outcome variables H (Only), O (Only), HO, and HT thus included a smaller subset of predictors than all other outcome variables. This variable selection means that we are unable to make certain comparisons between concepts involving H (Only), O (Only), HO, and HT.

**Table 4. Variable definitions.**

| Binary Outcome Variable | Variable Definition | Count TRUE | % TRUE |
|---|---|---|---|
| H (all) | All comments labeled hateful. | 148 | 17.98% |
| O (all) | All comments labeled offensive. | 281 | 34.14% |
| T (all) | All comments labeled toxic. | 288 | 35.00% |
| H (only) | Comments labeled only hateful, but not offensive or toxic. | 21 | 2.55% |
| O (only) | Comments labeled only offensive, but not hateful or toxic. | 59 | 7.17% |
| T (only) | Comments labeled only toxic, but not hateful or offensive. | 50 | 6.08% |
| HO | Comments labeled both hateful and offensive, but not toxic. | 13 | 1.58% |
| OT | Comments labeled both offensive and toxic, but not hateful. | 117 | 14.22% |
| HT | Comments labeled both hateful and toxic, but not offensive. | 15 | 1.82% |
| HOT | Comments labeled as all three of hateful, offensive, and toxic. | 99 | 12.03% |
| Overlap (any) | Comments labeled as any of hateful, offensive, and toxic. | 244 | 29.65% |

*Table Notes: The left-most column indicates binary outcome variable names. The center column defines each outcome variable. The two rightmost columns indicate how many comments and the percentage comments out of 823 total directed comments received a TRUE value for the given outcome variable.*

### *3.2 Venn Diagram & Mutual Information Ratio*

In addition to content analysis, we used the Venn diagram to visualize how much overlap exists between the sets of comments that received majority vote labels of H, O, and T (Venn, 1880). The Venn diagram visually shows the relations between categorical variables. Since we are interested in knowing how often annotators label comments as "H but not O" or "T but not H" (categorical variables), for instance, the Venn diagram is an appropriate analytical method for our study. If one concept is a subset of another, we expect that one circle fully absorbs another. The Venn diagram reveals the overarching patterns of overlaps between concepts.

Inspired by the idea of a performance ratio metric proposed in (Resnick et al., 2021), we also calculated a mutual information ratio (MIR) metric to quantify the directional relation between concepts. MIR quantifies how much information one worker's concept label provides to predict another worker's label of the *same* concept and of a *different* concept. Unlike the content analysis and Venn diagram relying on the majority vote labels, MIR operates on individual labels from every annotator. MIR is a directional measure of using a source concept ($S$) to approximate a target concept ($T$), while accounting for the inter-rater reliability of both concepts. Specifically, MIR computes the ratio between the amount of information gained (IG) – an information theoretic measure established by Shannon (Shannon, 1948) – about a random variable T given that a label of S is observed, and the IG about T given another label of T is observed. The MIR of concept T over concept S can be formally expressed as:

$$MIR(T|S) = \frac{IG(T, S)}{IG(T, T)}$$

where IG is the difference between the base entropy of T (i.e., H(T)) – or the amount of information needed to describe a random variable T – and conditional entropy of T given S (i.e., H(T|S)) – or amount of information needed to describe a random variable T given that a random variable S is observed. IG, base entropy, and conditional entropy can be formally expressed as:

$$IG(T, S) = H(T) - H(T|S)$$

$$H(T) = -\sum_{t \in \{0, 1\}} Pr(t) \, log(Pr(t))$$

$$H(T|S) = -\sum_{s \in \{0, 1\}, t \in \{0, 1\}} Pr(s, t) \, log(Pr(t|s))$$

If S does not contain any information about T, then H(T|S) = H(T) and IG(T, S) = 0. In most circumstances, a *T* label will be more useful than an *S* label in predicting another *T* label. Thus, the typical range of MIR is [0, 1]. A score of MIR = 0 indicates that the input label *S* contains zero information about the target T. A score of 1 indicates that the input label *S* can provide as much information as another random *T* label. A score of 0.9 indicates that we can get 90% of the information that a single *T* label provides about another *T* label by using a single *S* label instead. In rare cases, MIR can be greater than 1. For example, if *S* taps into exactly the same concept as *T*, but is providing a less noisy signal (e.g., resulted from a more competent annotator pool), then an *S* label will be more informative about a random *T* label than another *T* label would be, and the MIR can be greater than 1. MIR thus allowed us to understand the directionality of overlap between concept pairs. See Appendix for additional detail on how we calculated MIR.

## 4 Results

Our analyses suggest that hateful (H), offensive (O), and toxic (T) share some features and also have some unique features; H is a distinct concept relative to O and T; most T comments are also O, but most O comments are not also T; and *rater effect* significantly predicts labels. Subsections will refer to Table 5, which represents regression model outputs for each of our eleven outcome variables.

How We Define Harm Impacts Data Annotations**Table 5. Statistically significant predictors for directed comments in odds ratios.**

|  | Dependent variable: | | | | | | | | | | |
|---|---|---|---|---|---|---|---|---|---|---|---|
|  | H (all) | O (all) | T (all) | H (only) | O (only) | T (only) | HO | HT | OT | HOT | Overlap (any) |
| **Communication Style** | | | | | | | | | | | |
| Assertive | 0.857*** | 0.736*** | 0.820*** |  | 0.946** | 1.051** |  |  | 0.891*** | 0.885*** | 0.768*** |
| Passive | 0.820** | 0.691*** | 0.957 |  | 0.927 | 1.342*** |  |  | 0.857* | 0.858* | 0.713*** |
| Passive-Aggressive | 0.900*** | 0.776*** | 0.842*** |  | 0.987 | 1.066*** |  |  | 0.865*** | 0.918*** | 0.783*** |
| **Sarcasm** | 0.975 | 0.928 | 1.188*** | 0.972 | 0.940* | 1.255*** |  |  | 0.973 | 1.004 | 0.961 |
| **Pronoun Use** | | | | | | | | | | | |
| 1st Person | 0.884 | 1.007 | 0.788 |  |  | 0.976 | 0.912 |  | 0.962 | 0.943 | 0.833 |
| 2nd or 3rd Person | 0.944 | 1.046 | 0.999 |  |  | 1.042 |  | 0.913** | 1.045 | 1.002 | 0.970 |
| **Directed at** | | | | | | | | | | | |
| Group or Individual | 1.335* | 1.248 | 1.530** |  |  | 1.133 |  |  | 0.989 | 1.374** | 1.333 |
| Behavior | 1.229 | 1.205 | 1.392* |  |  | 1.099 |  |  | 0.981 | 1.339** | 1.243 |
| **Behavior** | | | | | | | | | | | |
| Accusation | 1.181*** | 1.004 | 1.084* |  |  | 1.014 |  |  | 0.922** | 1.137*** | 1.072* |
| Benign | 1.197 | 1.025 | 0.993 |  |  | 0.883* |  |  | 0.957 | 1.167 | 1.139 |
| Threat | 1.154** | 1.120 | 1.411*** |  |  | 1.171*** |  |  | 1.041 | 1.142** | 1.195** |
| Insult | 1.137*** | 1.127** | 1.151*** |  |  | 0.999 |  |  | 1.019 | 1.118*** | 1.158*** |
| Misinformation | 1.129 | 1.069 | 1.032 |  |  | 0.949 |  |  | 0.985 | 1.072 | 1.063 |
| Name-calling | 1.277*** | 1.313*** | 1.368*** |  |  | 1.013 | 1.021** | 1.026* | 1.093** | 1.198*** | 1.376*** |
| **Rater Effect** | | | | | | | | | | | |
| H | 5.750*** |  |  |  |  |  |  |  |  | 2.633*** | 1.730 |
| O |  | 4.690*** |  |  |  |  |  |  | 1.922*** | 1.073 | 2.249** |
| T |  |  | 7.298*** |  |  | 1.709*** |  | 1.141 | 1.274 | 1.518* | 2.076** |
| Constant | 0.626** | 0.699 | 0.474*** | 1.028*** | 1.094*** | 0.737** | 1.009* | 1.058 | 0.927 | 0.551*** | 0.581** |
| Observations | 823 | 823 | 823 | 823 | 823 | 823 | 823 | 823 | 823 | 823 | 823 |
| Log Likelihood | −319.061 | −442.662 | −448.148 | 355.090 | −45.276 | 91.732 | 549.717 | 496.403 | −251.955 | −192.227 | −401.520 |
| Akaike Inf. Crit. | 670.122 | 917.323 | 928.296 | −678.179 | 122.552 | −151.465 | −1,065.435 | −958.805 | 537.909 | 420.454 | 839.041 |

Note: *p<0.1; **p<0.05; ***p<0.01

*Table Notes: For binary variable "sarcasm", the holdout category is "n" or no sarcasm present. For categorical variable "communication style", the holdout category is "aggressive".*

## *4.1 Hateful is a Distinct Concept*

Venn diagrams (see Figure 1) show that comments were most frequently labeled as both T and O, and comments with H labels less frequently also received an O or T label. This finding suggests that H is a distinct concept relative to O and T. Represented in Table 6, low mutual information ratios (MIRs) for all concept pairs including H confirms our Venn diagrams' suggestion that H is a distinct concept relative to O and T – information about one does not provide much information about the other.

**Figure 1. Venn diagram showing overlap among H, O, and T.**

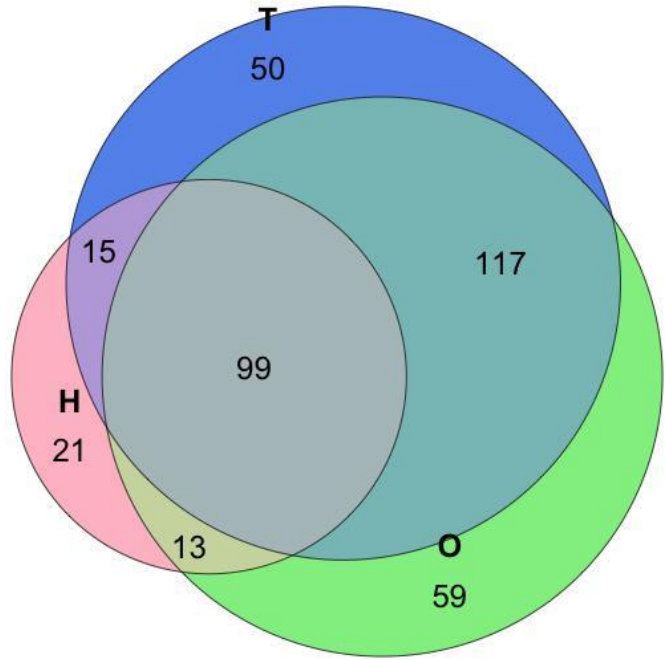

*Figure Notes: Numbers denote how many of the comments received the given H, O, or T label.*

**Table 6. Mutual information ratios between concept labels.**

|  | Target Concept | | |
| --- | --- | --- | --- |
| *Source Concept* | **Hateful** | **Offensive** | **Toxic** |
| **Hateful** | 1 | 0.292 (+/- 0.065) | 0.348 (+/- 0.084) |
| **Offensive** | 0.422 (+/- 0.102) | 1 | 0.741 (+/- 0.132) |
| **Toxic** | 0.413 (+/- 0.105) | 0.613 (+/- 0.099) | 1 |

*Table Notes: Values in parentheses denote the margin of error for each source->target concept pair.*

> H is the most distinct concept of the three concepts analyzed.

### *4.2 Toxic Communication Styles and Behaviors Resemble Offensive*

Confirming general patterns identified using Venn diagrams (Figure 1), higher mutual information ratios (MIRs) between O and T suggest that these concepts overlap to a greater degree with one another than with H (Table 6). This finding suggests that T and O are more closely related to each other than to H. However, MIR magnitudes tell us that we can get more information about a T label from an O label than we can get about an O label from a T label. These magnitudes suggest that while O and T are closely related, the ways in which they relate are not necessarily the same: T looks a lot like O, but O looks like T less often.

Reviewing the subset of comments labeled both O and T but not H, we see that communication styles (i.e., *aggressive*, etc.) and *accusation* confirm the Venn diagram and MIR findings that T provides more information about O than O provides about T. Specifically, by looking at directional resemblance of OT's odds ratios to O and T odds ratios (see Table 5), we find that: OT directionally resembles O across communication styles, *sarcasm*, and being *directed at a behavior, group, or individual*; and OT directionally resembles T in *pronoun use* and featured behaviors except for *accusation*, where it directionally resembles O. Of these predictors for OT, only the three communication styles, *accusation*, and *name-calling* are significant. Given that the majority of these significant predictors (four of five) directionally resemble O, those T comments represented in the OT subset (44.71% of all T comments, see Figure 1) resemble O comments. This resemblance could explain why MIRs suggest that we can get more information about a T label from an O label than the other way around.

> T *communication styles* and *behaviors* often resemble O's, but not the other way around.

### *4.3 Hateful is Hateful, Offensive is Aggressive, Toxic is Sarcastic and Threatening*

When looking at the subsets of H, O, and T that include comments labeled with *only* H, O, or T, we observe no overlapping characteristics between concepts. We find no significant predictors of comments labeled with only H. Comments labeled with only O are less likely to be assertive than aggressive, and less likely to be sarcastic. Comments labeled with only T are more likely to be assertive, passive, and passive-aggressive than aggressive; sarcastic; and feature threats, and are *less* likely to be benign. These analyses suggest that H, O, and T have some distinct characteristics.

> O is more likely to be *aggressive* and less likely to be *sarcastic* and feature *accusations*; T is less likely to be *aggressive* or *benign*, and more likely to be *sarcastic* and feature *threats*.

### *4.4 Hateful, Offensive, & Toxic are All Insulting, Call Names, and Affected by Who is Labeling*

From our analysis of *all* comments labeled as H, O, or T, we observe both characteristic differences and similarities between the three concepts. All three concepts are more likely to include insult and name-calling and be affected by raters' individual interpretive lenses (i.e. *rater effect* predictor). At the same time, T is more likely to be sarcastic and be directed at behaviors, and H and T are both more likely to feature accusation or threat and be directed at a group or individual. We observe fewer characteristic differences for O, and that there are more shared characteristics between H and T than between H and O. While comments labeled with *only* H, O, or T reveal that the three concepts have some distinct characteristics, looking at *all* comments labeled with H, O, or T reveals that the three concepts also share some characteristics.

> H, O, and T are all likely to include *insults* and *name-calling* and to be predicted by *rater effect*.

### *4.5 Who Labels is the Strongest Predictor for Comments' Labels*

In model results with outcome variables referring to all comments labeled with H, O, or T, *rater effects* are statistically significant and exhibit the greatest magnitude among predictors for H, O, and T. However, we find that for comments featuring any overlapping concepts, O and T *rater effects* are significant. For OT comments, O *rater effect* is significant. For HOT comments, H and T *rater effects* are significant. These subset analyses suggest that *rater effect* is most commonly a significant predictor for T in subsets, followed by O then H. However, whether statistically significant or not, *rater effect* has the highest magnitude effect on all outcome variables.

> *Rater effect* has the highest predictive effect for all comments labeled as H, O, or T.

## 5 Discussion

Annotators labeled content from social media as H, O, and/or T. We analyzed both the annotations and the comments and found four key findings about these labels and annotation tasks:

1. H is the most distinct concept of the three concepts analyzed;
2. T *communication styles* and *behaviors* often resemble O's, but not the other way around;
3. O is more likely to be *aggressive* and less likely to be *sarcastic* and feature *accusations*; T is less likely to be *aggressive* or *benign*, and more likely to be *sarcastic* and feature *threats*;

4. H, O, and T are all likely to include *insults* and *name-calling*, and to be predicted by *rater effect*;
5. *Rater effect* has the highest predictive effect for all comments labeled as H, O, or T.

We unpack explanations for these findings and related recommendations in the following subsections. Specifically, we discuss how the specificity (or non-specificity) of a harm concept's definition may explain some of our findings, guidance for developing specificity in concept definitions, and suggestions for future research that investigates how features beyond concept definitions like annotators' individual characteristics (rater effect) can affect annotators' interpretations of harm concepts.

## *5.1 Definitional Specificity*

It is possible that some of our findings can be explained by the specificity of the concept definitions that we provided to annotators. Recall that the definition of H we provided to annotators was the only one to specify comment features (see Table 1 in section 3 for reminder of definitions). O's definition referred to the content of comments – "obscene" – but did not specify what "obscene" content means. T's definition, most of O's definition, and some of H's definition ask the annotator to infer the comment author's intent or the comment's possible effects on readers, neither of which annotators had direct information about. Thus, it is possible that annotators interpreted H as the most distinct concept because they were able to more clearly identify H features using the comment-related information they had and H's comment-oriented definition. In contrast, it is possible that annotators were less able to distinguish O and T concepts because they did not have comment author and reader information necessary to make the inferences that O and T definitions required.

Related, it is possible that *rater effect* is significant for H less frequently across outcome variables than for O and T because H's definition is the most interpretable given the information we provided annotators, potentially reducing the effects of individual rater characteristics. Conversely, it is possible that *rater effect* is most often significant for T relative to the other two concepts because T's definition entirely refers to author intent or the effect of a comment on a reader. Annotators had direct information about neither of these features in Wu et al.'s data annotation task (2023), potentially making individual rater's interpretive differences more impactful.

This explanation for some of our findings highlights the potential importance of future researchers preparing annotation tasks that specifically consider *what* researchers are asking annotators to label (i.e., comment features, author intentions, effect of a comment on a reader), and *what information they provide to annotators* to be able to complete this task. The more specific an annotation task is and the closer the match between what the task asks annotators to do and the information they are provided, the less impactful an annotator's individual interpretive lens may be on the labeled outcome.

## *5.2 Non-Specific Definitions May Still Drive Conceptual Difference*

In some instances, the definitions of H, O, and T that Wu et al.'s (2023) annotation task used specified the types of behaviors and tone (i.e., *communication style* and *sarcasm* features) that annotators could look for to identify the presence of each concept. However, these definitions did not further specify what constitutes each behavior or tone. For example, while O's definition specifies "hurtful, derogatory or obscene" comments as characteristics of O, it did not define what it meant by these characteristics. It is possible that *aggressiveness,* not being *sarcastic*, and not featuring *accusations* describes how annotators interpreted "hurtful, derogatory or obscene" for O. Similarly, it is possible that *threat* describes how annotators interpreted and applied the component of T's definition that referred to "likely to make readers want to leave a discussion", and that *sarcastic* reflects annotators' interpretation and application of "rude, disrespectful, or unreasonable" specified in T's definition. If it is the case that *communication style*, *behavior*, and *sarcasm* are the most significant predictors of a comment being labeled as any of H, O, or T, these differences between concepts could be explained by differences in concept definitions. However, since Wu et al.'s annotation task did not ask annotators specifically about how they interpreted and applied the behaviors and tones mentioned in these concept definitions, it is possible that this is not the case. Future research should assess this hypothesis in greater depth.

## *5.3 Resource Constraints on Specificity*

While our findings encourage researchers to clearly define their annotation tasks and provide matching information to annotators to be able to complete the specified task, we also recognize that resources constrain how much agency researchers actually have in making these ideal decisions. For example, a small research lab may have a research goal of detecting the presence of a particular definition of harm in an online gaming community. Their definition of harm may differ from those of mainstream harmful content detection algorithms. The lab would ideally like to train a new algorithm to detect this particular type of harmful content, but is met with severe resource constraints: it does not have access to server systems able to train such a harmful content detection algorithm. Despite this mismatch in off-the-shelf algorithms' ability to answer the lab's research question, the lab's resource constraints outweigh their ability to design a new detection algorithm trained on datasets annotated according to specific instructions, and the lab opts to use the off-the-shelf algorithm. In this situation, we propose two steps that can be taken to (a) ensure that researchers adequately report on the constraints of their findings and (b) increase researchers' ability to adapt existing off-the-shelf algorithms to match researchers' needs in resource-constrained environments.

Where the concept definitions a researcher wants to analyze differ from available harmful content detection resources, providing sufficient background to contextualize contributions relative to harmful content detection instruments used becomes all the more important. For example, rather than imagining that findings about "hateful" content, as defined by Davidson et al. (2017), apply to all harmful content in all communities, a researcher can clearly articulate definitional and community-specific contexts to which their findings apply. While field-specific conventions may prioritize generalizability, appropriately contextualizing contributions can help to ensure that findings are not inappropriately generalized or ineffectively used to address a potentially unrelated harmful content detection issue.

Beyond contextualization, we also encourage researchers to consider how they might adapt available resources like Perspective API or Open AI's Moderation API to address their research goals. For example, in addition to toxicity scores for content, Perspective API returns attributes "insult", "identity attack", "profanity", and "threat" as per their definitions of these terms.[2] It is not possible for a researcher external to Perspective API's development team to change the weights of these attributes when using Perspective API to label new datasets. However, a researcher could consider developing their own dataset trained on a content sample labeled by Perspective API, but with differently weighted attributes. Using this approach, a researcher could cobble together a weighted definition of harm that more closely matches their research needs.

Finally, we encourage better resourced institutions like Jigsaw or Open AI that generously make their harmful content detection algorithms like Perspective API available for public use, to consider how to make their resources adaptable to specific harmful content detection needs. For example, these organizations could consider soliciting requests from their API users about which additional harm-related attributes would be helpful to collect labels for so that researchers can more easily construct bespoke models with attributes that match their harm-detection needs. Much like validated survey instruments can be adapted to some degree for contextually-specific research goals, we encourage these organizations to consider: how might harmful content detection algorithms be made adaptable for contextually-specific harmful content detection moderation and research goals?

## *5.4 Beyond Definitions*

In addition to differences in how the data were labeled driven by the definitions of H, O, and T provided to annotators, we also identify three features not encompassed by the definitions that may have driven annotators' labeling: insults, name-calling, and rater effect. In this section we articulate future research to further investigate these features and how they relate to annotation task design.

### 5.4.1 Insults and Name-Calling

While there may not seem to be much overlap between H, O, and T as concepts definitionally, annotators show that insults and name-calling are common features across all three concepts operationally. It is possible that these common features reflect how annotators' perceive harm in general, and since each of H, O, and T refer to variations of harm, annotators tended to look for insults and name-calling as identifying features of each concept even though we did not specifically ask them to. In order to assess whether this explanation is accurate, we encourage future research to study how annotators perceive and define "harm" when they review social media comments when they are not provided with a specific definition of harm in advance.

---

[2] See https://developers.perspectiveapi.com/s/about-the-api-attributes-and-languages?language=en_US for definitions of these attributes.

**5.4.2 Annotators' Individual Characteristics (Rater Effect)**

Finally, the predictor with the highest significant effect size across H, O, and T concepts is *rater effect*. This finding opens important opportunities for future work to investigate *how* individual annotators' perspectives – and consequently groups of annotators' perspectives – affect annotations. Would similar annotations result from an annotator pool that is majority gender non-binary rather than majority male or female, for example? What if annotators were labeling under different labor conditions – fixed salary and enforced breaks, for instance? We encourage future research to investigate these questions.

## 6 Conclusion

Our analyses reveal that annotators do not interchangeably use the concepts hateful, offensive, and toxic. We show that the more specific a definition of harm is in articulating which comment features to identify, the less annotators' individual characteristics affect their labeling decisions. When harm definitions ask annotators to identify ambiguously defined comment features or features that annotators do not have information about (e.g., comment author's intent), annotators' individual characteristics have a greater effect on labeling decisions. These findings highlight the importance of specifically defining harm in annotation tasks and providing annotators with the information they need to complete the requested task.

We recommend that researchers make careful, specific choices about which concepts to analyze, how to define them, and how to provide annotators with adequate information to assess concepts. However, recognizing that researchers are often resource constrained and cannot always choose how to define harm concepts, we encourage them to appropriately contextualize their findings when concepts of interest may differ from off-the-shelf algorithms' harm definitions. We also encourage harmful content detection algorithm providers to consider how to ensure their instruments can adapt to contextually-specific harmful content detection goals. Finally, we propose that future research investigate how and which annotators' individual characteristics affect their interpretation of harm concepts. Our research shows that how we define harm for annotation tasks is integral to developing contextually-specific content moderation support algorithms that mitigate people's experiences of harm in online communities.

## Acknowledgements

We are thankful for our colleagues' help and feedback on this project. Morgan Wofford provided comments on our content analysis process. Corey Powell and Abner Heredia at Consulting for Statistics, Computing, and Analytics Research (CSCAR) at the University of Michigan advised on statistical analyses. Shubham Atreja, Gina Brandolino, Lizhou (Leo) Fan, Allegra Fonda-Bonardi, Ji Eun Kim, Lingyao Li, and Allie Piippo provided comments on earlier drafts. Nico Wilkins provided copyediting services funded by the UMSI Carnegie Fund for faculty development.

## Funding

The author(s) disclosed receipt of the following financial support for the research, authorship, and/or publication of this article: This material is based upon work supported by the National Science Foundation under Grant No. 1928434.

# 7 Appendix

## 7.1 MIR Computation Details and Pseudocode

Cross-replication reliability (xRR) motivates MIR design (Wong et al., 2021). Applying principles of xRR to our context, we considered random draws from existing label sets as independent replications and computed the average information gain across replications. This bootstrapping method allows us to compute a metric IG(T, T) that captures internal annotation reliability within the same concept *T*. Using just the majority voting label of *T*, IG(T, T) cannot be computed. We provide the pseudocode of MIR computation below.

**Input:** *num_bootstrap*, *num_item*, source_name, target_name
**Output:** one numerator and one denominator confusion matrix
**While** *num_bootstrap* not equals to zero, **do**
   Randomly sample *num_item* rows, without replacement, from the original dataframe;
   **For** each row in the sampled dataframe, **do**
     Randomly sample two labels ***T1*** and ***T2***, without replacement, from row[target];
     Use T2 as the target label

     When source and target are different:
       Randomly sample one label ***S*** as the source;
     When source and target are the same:
       Use T1 as S;

     Generate numerator confusion matrix with y_true = ***S*** and y_pred = ***T2***;
     Generate denominator confusion matrix with y_true = ***T1*** and y_pred = ***T2***;
   **End**
   *num_bootstrap* -= 1
**End**

We also considered logistic regression (LR) to measure this phenomenon, but while LR measures the correlation between two concepts, it does not integrate internal annotation uncertainty for each concept. Inter-rater reliability (IRR) addresses some of this concern and is designed to quantify the annotation uncertainty among different independent raters for one concept. However, since we were interested in understanding uncertainty *between two* concepts, IRR was not an appropriate analysis method for our study either.

# How We Define Harm Impacts Data Annotations

## 7.2 Likelihood Ratio Tests

**Table 7. Likelihood ratio test outputs for each model.**

| Model | #Df | LogLik | Df | Chisq | Pr(>Chisq) |
|---|---|---|---|---|---|
| H (All) | 17 | -318.06 | | | |
| Null | 2 | -380.18 | -15 | 124.24 | 0.0000 |
| O (All) | 17 | -441.66 | | | |
| Null | 2 | -558.48 | -15 | 233.65 | 0.0000 |
| T (All) | 17 | -447.15 | | | |
| Null | 2 | -553.71 | -15 | 213.12 | 0.0000 |
| H (Only) | 3 | 353.48 | | | |
| Null | 2 | 352.41 | -1 | 2.13 | 0.1443 |
| O (Only) | 6 | -47.85 | | | |
| Null | 2 | -52.70 | -4 | 9.71 | 0.0456 |
| T (Only) | 17 | 92.73 | | | |
| Null | 2 | 10.59 | -15 | 164.29 | 0.0000 |
| HO | 3 | 548.23 | | | |
| Null | 2 | 545.67 | -1 | 5.12 | 0.0237 |
| HT | 6 | 493.51 | | | |
| Null | 2 | 487.80 | -4 | 11.41 | 0.0223 |
| OT | 18 | -250.95 | | | |
| Null | 2 | -301.94 | -16 | 101.97 | 0.0000 |
| HOT | 19 | -191.23 | | | |
| Null | 2 | -243.56 | -17 | 104.66 | 0.0000 |
| Overlap (Any) | 19 | -400.52 | | | |
| Null | 2 | -522.78 | -17 | 244.53 | 0.0000 |

## 7.2 Sample Coded Comment

**Table 8. Sample comment with codes assigned from codebook.**

**Comment:** "@ksatnews Good for him. Glad he caught it. I do wish him well though. Maybe he will think again about sending kids to school without wearing a mask."

| Feature | Code |
|---|---|
| **Directedness** | Directed |
| Directed at | Individual |
| **Pronoun Use** | 3rd person |
| **Behavior** | Threat |
| **Multiple Behaviors** | No |
| **Communication Style** | Aggressive |
| **Sarcasm** | No |

## 7.3 Sample HOT Comments

**Table 9. Sample comments annotated as any one of, or any combination of H, O, or T.**

| Concept | Sample Comment |
|---|---|
| H | @Reuters This is a genius plan to convince anti-vaxxers to get bitten by vipers. |
| O | @TheLeadCNN @BillKristol How the hell are you CNN people able to stomach this horse crap. |
| T | @globalnews @GlobalCalgary Ok Canada here's out time to kick that lying disrespected cheating no go leader out of power! So pleeeeeeeease do the right thing and don't vote liberal cause all he will do and it will happen is ruin our country and make all other countries laugh and treat us like him garbage! |
| HO | @JamesGRickards all of our major cities that are run by the Dims are cesspools and very dangerous. I'll stick to the road less traveled and the countryside, real America. https://t.co/wGtep8FIVN |
| HT | @AliVelshi They blame the people coming crossing the Border, and that didn't work, so they start on the blacks. The Republican blame game. |
| OT | @Jedi_MAGA @sillyfools09 @thehill No one is killing unwanted babies. It's a blob of cells, a parasite. |
| HOT | ALL cops that where at the capital need a kill switch....tell their truth before they get offed. |